# Gabor Filter and Rough Clustering Based Edge Detection

Chandranath Adak
Department of Computer Science and Engineering
University of Kalyani
West Bengal-741235, India
mr.c.adak@ieee.org

*Abstract*—This paper introduces an efficient edge detection method based on Gabor filter and rough clustering. The input image is smoothed by Gabor function, and the concept of rough clustering is used to focus on edge detection with soft computational approach. Hysteresis thresholding is used to get the actual output, i.e. edges of the input image. To show the effectiveness, the proposed technique is compared with some other edge detection methods.

## I. INTRODUCTION

*Edge detection* [1-6] becomes one of the challenging issue regarding image processing (more specifically, image segmentation) for the last three decades. It is the process to identify the border-line or boundary between a pair of objects/regions. A sufficient number of computer vision and pattern recognition techniques are dependent on edge detection as a priori (pre-processing) stage. An accurate and efficient edge-detector increases the performance of different applications related to image processing, pattern recognition, machine vision (with artificial intelligence) problems, e.g. object-based coding [7], image segmentation [8-9], image retrieval [10] etc.

This edge detection problem can be viewed as a clustering [11] process where the task is to classify the data into two sets: *edge* and *non-edge*. The patterns, 'within the cluster' and 'between the clusters' are *homogeneous* and *heterogeneous* respectively.

Here Gabor filter [12-13] is used to smooth the image. Rough clustering [14, 31] with rough set and Pawlak's accuracy [15] is used to modify the production of nonmaxima-suppressed image [16]. Hysteresis thresholding [17] is used to produce final output image with detected edges.

## II. GABOR FILTER

A *Gabor filter* (Dennis Gabor, 1946) is a linear filter whose impulse response is the multiplication of a harmonic function with a Gaussian function [18-20]. As per *convolution theorem*, the convolution of Fourier Transformation (FT) of harmonic function and FT of Gaussian function is nothing but FT of a Gabor filter's impulse response [ $FT(Gabor) =$ $FT(Harmonic)$ ★ $FT(Gaussian)$ ]. The filter consists of a real and an imaginary component, which represent the orthogonal directions. The two components are used individually or in a complex form.

*Complex* :

$g(x,y;\lambda,\theta,\varphi,\sigma,\gamma) = exp(-(x_1^2 + \gamma^2 y_1^2)/2\sigma^2) \cdot exp(i.(2\pi x_1/\lambda + \varphi))$
---------- (1)

*Real* :

$g(x,y;\lambda,\theta,\varphi,\sigma,\gamma) = exp(-(x_1^2 + \gamma^2 y_1^2)/2\sigma^2) \cdot cos(2\pi x_1/\lambda + \varphi)$
---------- (2)

*Imaginary* :

$g(x,y;\lambda,\theta,\varphi,\sigma,\gamma) = exp(-(x_1^2 + \gamma^2 y_1^2)/2\sigma^2) \cdot sin(2\pi x_1/\lambda + \varphi)$
---------- (3)

Where, $x_1 = xcos\theta + ysin\theta$ and $y_1 = -xsin\theta + ycos\theta$

In *eq.-1,2,3* :
 $\lambda$ : wavelength of sinusoidal factor,
 $\theta$ : orientation of normal to parallel stripes,
 $\varphi$ : phase offset,
 $\sigma$ : sigma of Gaussian envelope,
 $\gamma$ : spatial aspect ratio (specifies the ellipticity).

Daugman (*J. Daugman; 1980, 1985*) extended the Gabor filter into two dimensions [12].

## III. ROUGH CLUSTERING

*Rough clustering* (Prado, Engel, Filho, 2002; Voges, Pope, Brown, 2002) is an expansion work of rough (approximation) sets, which is pioneered by Pawlak (1982, 1991) [21].

### A. Information System Framework

In Rough set theory, an assumption is granted, i.e. information is related with each and every entry of the data matrix. The over-all information expresses the completely available object-knowledge. More precisely, the information system is a pair of tuples, $S=(U,A)$, where U is a non-empty finite object set called as universe and $A=\{a_1,...,a_j\}$ is a non-empty finite attribute set on *U*. With every attribute $a \in A$, a set

$V_a$ is allied such that $a : U \rightarrow V_a$. The set $V_a$ is called the domain (value) set of $a$.

### B. Equivalence Relation

An associated *equivalence relation* resides with any $P \subseteq A$,

$$IND(P) = \{(x,y) \in U^2 \mid \forall a \in P, a(x)=a(y)\}.$$

The equivalence relation $IND(P)$ is termed as a *P-indiscernibility* relation. The partition of $U$ is a family of all *equivalence classes of IND(P)*, denoted by $U/IND(P)$ or, $U/P$. Those $x$ and $y$ are *indiscernible* attributes from $P$, when $(x,y) \in IND(P)$.

### C. Rough Set

The main thing of rough set is the equivalence between objects (known as indiscernibility). The equivalence relation is formed with same knowledge-based objects of the information system. The partitions (formed by division of equivalence relations) build the new subsets. An information system $S=(U,A)$ is assumed, such that $P \subseteq A$ and $X \subseteq U$. The subset $X$ (using information contained in attributes from $P$) is described by constructing two subsets: *P-lower* approximations of $X$ ($P_*(X)$) and *P-upper* approximations of $X$ ($P^*(X)$), where:

$$P_*(X) = \{ x \mid [x]_P \subseteq X \} \text{ and } P^*(X) = \{ x \mid [x]_P \cap X \neq \emptyset \}.$$

Sometimes, an additional set $(P_D(X))$, i.e. the difference between the upper approximation $(P^*(X))$ and the lower approximation $(P_*(X))$ becomes very effective in analysis.

$$P_D(X) = P^*(X) - P_*(X)$$

The *accuracy* $(\alpha_P)$ of the rough-set (Pawlak, 1991) representation of the set $X$ is as follows:

$$0 \leq ( \alpha_P (X) = |P_*(X)| / |P^*(X)| ) \leq 1$$

Rough clustering is the extension of rough sets, containing two additional requirements: an ordered attribute value set and a distance measurement (Voges, Pope & Brown, 2002). As like standard clustering techniques, distance measurement is done by ordering value set, and clusters are generated by these distance measure.

## IV. PROPOSED TECHNIQUE

The steps of proposed technique are as follows:

Step 1: Computational overhead is reduced by transformation of each RGB image pixel value into a single valued attribute:
$Pixel_T = Pixel_R + 2*Pixel_G + 3*Pixel_B$ ---(4)

Step 2: Single valued transformed image pixel values ($Pixel_T$) are the input data set $X = \{x_1, x_2, ..., x_n\}$.

Step 3: Standard deviation $(\sigma)$ is calculated:

$$\sigma = \sqrt{\frac{1}{n}\sum_{i=1}^{n}(X_i - \bar{X})^2} \quad ---(5)$$

Where $n$ is the total number of pixels and $\bar{X}$ is the arithmetic mean of the values $X = \{x_1, x_2, ..., x_n\}$, defined as:

$$\bar{X} = (x_1 + x_2 + \cdots + x_n)/n \quad ---(6)$$

Step 4: $f(x,y)$ denotes the input image and $G(x,y)$ is the Gabor function (i.e. spatial domain sinusoidal modulated Gaussian) impulse response for the imaginary part as eq.3. A smoothed image $f_s(x,y)$ is formed by convolution of $G$ and $f$:

$$f_s(x,y) = G(x,y) \star f(x,y) \quad ---(7)$$

Step 5: The gradient magnitude $(M(x,y))$ and direction $(\alpha(x,y))$ [22] are calculated:

$$grad(f_s) \equiv \begin{bmatrix} g_x \\ g_y \end{bmatrix} = \begin{bmatrix} \frac{\partial f_s}{\partial x} \\ \frac{\partial f_s}{\partial y} \end{bmatrix} \quad ---(8.1)$$

$$M(x,y) = mag( grad(f_s) ) = \sqrt{g_x^2 + g_y^2} \quad ---(8.2)$$

and,

$$\alpha(x,y) = \tan^{-1}(g_y/g_x) \quad -----(9)$$

Step 6: Thinning: *non-maxima suppression modified with rough-clustering based Pawlak's accuracy*: $d_1, d_2, d_3, d_4$ denotes four basic edge directions - $0°, 45°, 90°, 135°$ for a 3X3 region centred at $p_5(x,y)$ with its eight-connected neighbours.

| $p_1(x-1, y-1)$ | $p_2(x-1, y)$ | $p_3(x-1, y+1)$ |
|---|---|---|
| $p_4(x, y-1)$ | **$p_5(x, y)$** | $p_6(x, y+1)$ |
| $p_7(x+1, y-1)$ | $p_8(x+1, y)$ | $p_9(x+1, y+1)$ |

Fig.1. Eight-connectivity of a pixel $p_5(x,y)$

The direction $d_k$, that is closest to $\alpha(x,y)$ is found.

Along this direction $d_k$, we calculate *P-lower* and *P-upper* approximations of $M(x,y)$ denoted as $P_*(M)$ and $P^*(M)$ : if the value of $M(x,y)$ is less than at least one of its two neighbours along $d_k$, then it belongs to $P_*(M)$, otherwise $P^*(M)$.

The *Pawlak's accuracy* $(\alpha_P)$ is calculated :
$$\alpha_P(M) = |P_*(M)| / |P^*(M)|$$

*if ($\alpha_P(M) < T$)*
  $g_N(x,y) = 0$
*else*
  $g_N(x,y) = M(x,y)$

$T$ is a threshold [22-23] value ( $0 \leq T \leq 1$ ) and $g_N(x,y)$ is the rough-clustering based nonmaxima-suppressed image.

Step 7: To detect the final edges, *hysteresis thresholding* [24] is used as *Canny edge detector* [1].

[ *used convention*: edge pixel color = white , non-edge and background pixel color = black ] .

## V. EXPERIMENTAL RESULTS AND COMPARISON

To assess the immovability and accurateness of the proposed technique, the results are obtained from different images and compared with other existing methods.

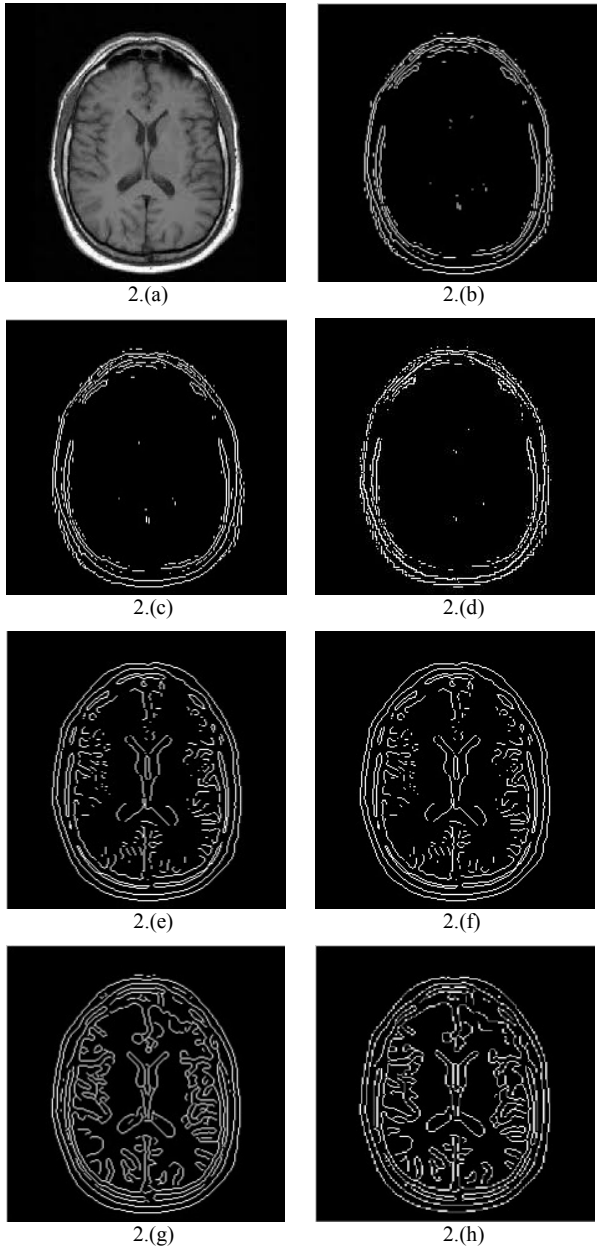

Fig.2. *Brain MRI*, compared with existing methods: (a) Original, (b) Sobel, (c) Prewitt, (d) Roberts, (e) Laplacian of Gaussian, (f) Zero-Cross, (g) Canny, (h) Proposed method (*T=0.5*).

In *fig.2*, the proposed method gives good result, it gives noise-free output comparably; it will be more clear in *fig.3*.

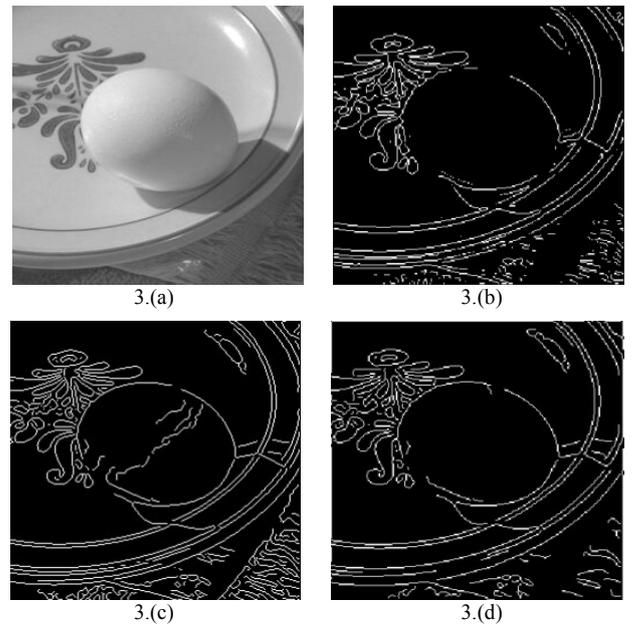

Fig.3. *Egg on Plate* [25]: (a) Original, (b) Laplacian of Gaussian, (c) Canny, (d) Proposed method (*T=0.55*).

In *fig.3.(c)*, the Canny-edge detector gives the noisy line on egg, but in *fig.3.(d)*, the proposed method removes the noisy edges.

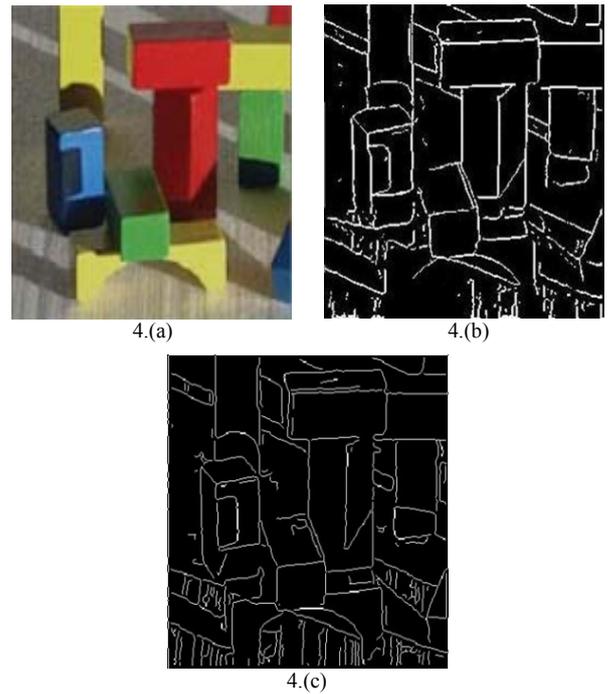

Fig.4. *Color Bars*: (a) Original, (b) Method reported in [26] and color is inverted, (c) Proposed method (*T=0.5*).

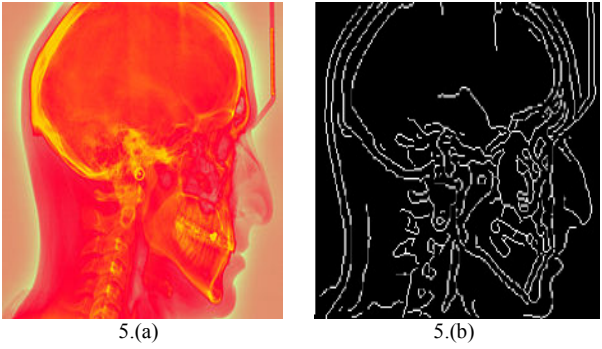

Fig.5. *Computed Radiography (CR)* [27]: (a) Original, (b) Proposed method (*T=0.6*).

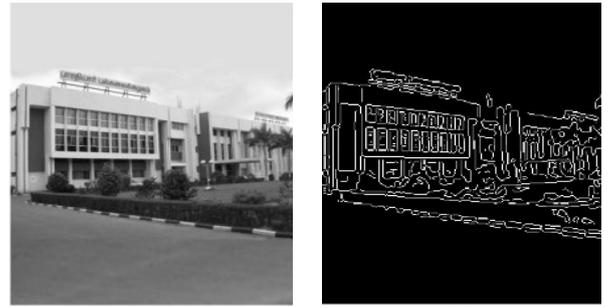

Fig.8. *Image of Bharathiar University, Coimbatore-641046, India,* comparisons reported in [30]: (a) Original, (b) Proposed method (*T=0.5*).

From *fig.4-8*, it is clearly shown that the proposed method gives clear noise-free edges with good continuity.

## VI. CONCLUSION AND FUTURE WORK

The proposed technique is tested on different images (e.g. *medical images*: MRI, CR, CT, X-Ray; *remotely sensed images*: islands, urban areas, country, planets, i.e. satellite images; *real life object images*: flower, egg, wheel, color bars, house, household things etc.). It produces stable, more noiseless and fairly good results in every case, which assesses the high robustness of this technique. The performance of the proposed method is compared with some classical edge detection techniques (e.g. Sobel, Prewitt, Roberts, Laplacian of Gaussian, Zero-Cross, Canny etc.) and methods reported in some papers [25-26, 29, 30]. Even though this method produces better results, it fails to image-shadow elimination [32], so the next venture will be definitely overcome this limitation and use genetic algorithm [33] to make the system more efficient.

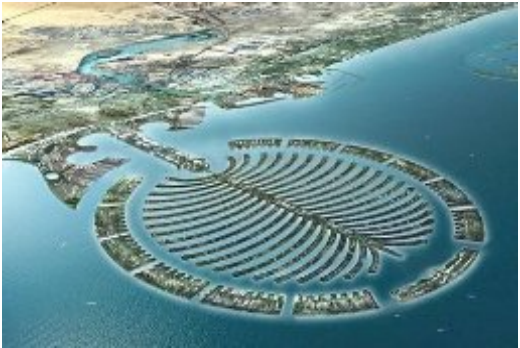

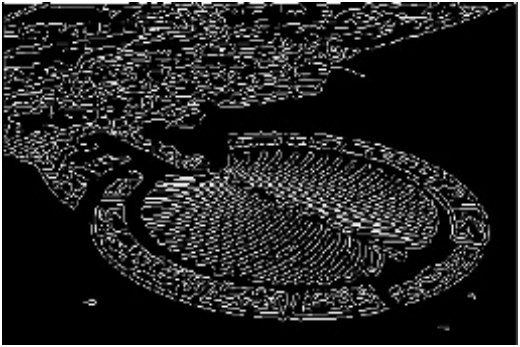

Fig.6. *Remotely Sensed Image: Palm Island, Dubai* [28]: (a) Original, (b) Proposed method (*T=0.58*).

### ACKNOWLEDGMENT

I would like to heartily thank *Prof. Bidyut B. Chaudhuri*, IEEE Fellow, Head, Computer Vision and Pattern Recognition Unit, Indian Statistical Institute, Kolkata 700108, India, for discussion various aspects of this research field.

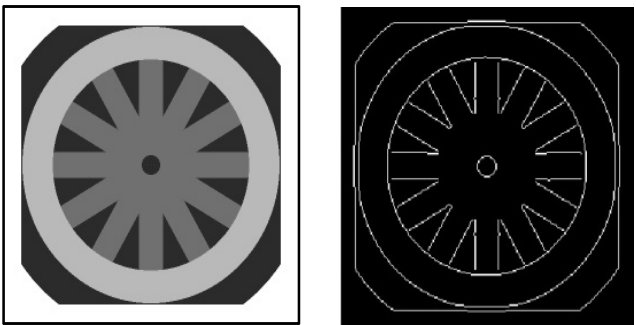

Fig.7. *Image of Wheel*, comparisons reported in [29]: (a) Original, (b) Proposed method (*T=0.57*).